\documentclass{article}
\usepackage{amsmath,graphicx,mlspconf}
\usepackage[table]{xcolor}  

\copyrightnotice{U.S.\ Government work not protected by U.S.\ copyright}

\copyrightnotice{979-8-3503-2411-2/25/\$31.00 {\copyright}2025 Crown}

\copyrightnotice{979-8-3503-2411-2/25/\$31.00 {\copyright}2025 European Union}

\copyrightnotice{979-8-3503-2411-2/25/\$31.00 {\copyright}2025 IEEE}

\toappear{2025 IEEE International Workshop on Machine Learning for Signal Processing, Aug.\ 31-- Sep.\ 3, 2025, Istanbul, Turkey}

\title{UTILIZING DYNAMIC SPARSITY ON PRETRAINED DETR}
\name{%
   Reza Sedghi$^{\star}$%
   \qquad Anand Subramoney$^{\dagger}$
   \qquad David Kappel$^{\star}$\thanks{RS is funded by BMBF project EVENTS (16ME0733). DK is funded by project SAIL (grant no. NW21-059A). The authors gratefully acknowledge the Gauss Centre for Supercomputing e.V. (www.gauss-centre.eu) for providing computing time on GCS JUWELS at Jülich Supercomputing Centre.}%
}
\address{%
   $^{\star}$ CITEC, Bielefeld University, Germany \\%
   $^{\dagger}$  Department of Computer Science, Royal Holloway, University of London, UK%
}

\begin{document}

\maketitle

\begin{abstract}

Efficient inference with transformer-based models remains a challenge, especially in vision tasks like object detection. We analyze the inherent sparsity in the MLP layers of DETR and introduce two methods to exploit it without retraining. First, we propose Static Indicator-Based Sparsification (SIBS), a heuristic method that predicts neuron inactivity based on fixed activation patterns. While simple, SIBS offers limited gains due to the input-dependent nature of sparsity. To address this, we introduce Micro-Gated Sparsification (MGS), a lightweight gating mechanism trained on top of a pretrained DETR. MGS predicts dynamic sparsity using a small linear layer and achieves up to 85–95\% activation sparsity. Experiments on the COCO dataset show that MGS maintains or even improves performance while significantly reducing computation. Our method offers a practical, input-adaptive approach to sparsification, enabling efficient deployment of pretrained vision transformers without full model retraining.
\end{abstract}

\section{Introduction}
Transformer-based architectures have achieved remarkable success across a wide range of domains, from natural language processing to computer vision \cite{islam2306comprehensive}. However, their substantial computational cost remains a major bottleneck for real-world deployment \cite{he2024matters}. To address this, a significant research trend has emerged around Mixture of Experts (MoE) models \cite{clark2022unified, geva2020transformer, shazeer2017outrageously}, which introduce conditional computation to activate only a subset of the model for each input. This approach allows for scaling up model capacity without a proportional increase in computational load, and has proven especially effective for large language models (LLMs) \cite{liu2024deepseek, wang2024scaling, shen2023mixture}.

\begin{figure}[!t]

\begin{minipage}[b]{1.0\linewidth}
  \centering
  \centerline{\includegraphics[width=8.5cm]{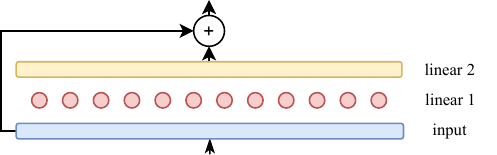}}
  \vspace{0.25cm}
  \vspace{0.25cm}
\end{minipage}
\begin{minipage}[b]{1.0\linewidth}
  \centering
  \centerline{\includegraphics[width=8.5cm]{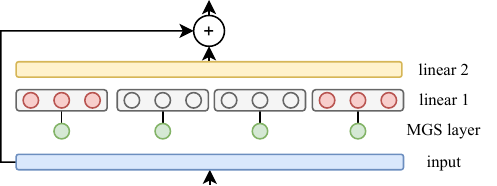}}
 \vspace{0.25cm}
\end{minipage}
\caption{(Top) Standard feedforward (FF) block used in DETR’s encoder and decoder layers. Blue represents the input hidden state, followed by a linear layer (gray), ReLU activation, and a second linear layer.
(bottom) Our proposed Micro-Gated Sparsification (MGS) structure. A lightweight gating layer (green) with sigmoid activation is added before the first linear layer to dynamically mask small groups of neurons based on input, reducing unnecessary computation at inference time.}
\label{fig:base}
\end{figure}
Despite these opportunities, most existing work on sparsification through Mixture of Experts (MoE) and related techniques has focused on extremely large-scale models, primarily in the domain of language modeling. Applications of MoE to vision transformers, and in particular to models like DETR \cite{carion2020end}, are far less explored. This may be due in part to the fact that DETR, while computationally intensive, does not reach the massive scale where MoE's benefits traditionally outweigh its complexity. Furthermore, integrating large-scale MoE architectures into DETR would require retraining the model from scratch, which is both computationally expensive and time-consuming given DETR's already costly training procedure \cite{zhu2021deformable}.

We investigate an alternative strategy: designing lightweight mechanisms that simulate expert-like conditional computation directly on top of pretrained models. Our goal is to make DETR more efficient by predicting and exploiting existing redundancy in its computations, rather than introducing heavy architectural modifications.
Specifically, we introduce two methods aimed at dynamically predicting and exploiting sparsity in the MLP (feedforward) layers of a pretrained DETR (ResNet-50) model:

\begin{itemize}
    \item A heuristic rule-based approach called Static Indicator-Based Sparsification (SIBS), which infers sparsity based on observed activation patterns.

    \item A learnable approach called Micro-Gated Sparsification (MGS), which trains a lightweight gating module to predict neuron group activations dynamically during inference.
\end{itemize}

Through detailed ablation studies and experiments on the COCO dataset, we demonstrate that significant redundancy exists in the feedforward (MLP) layers of DETR. Not only can this redundancy be exploited to achieve substantial computation savings, but in some cases, selectively masking out redundant neurons even leads to slight improvements in object detection performance. Our findings highlight the potential of fine-grained, input-adaptive sparsity prediction for practical efficiency gains in vision transformer models.





\section{background: related works}

Efficient deployment of transformer models for vision tasks, such as DETR, has become increasingly important as these models grow in size and complexity. DETR leverages transformer encoders and decoders for object detection. Several works have attempted to improve efficiency in DETR, primarily by focusing on the attention mechanisms. For instance, Deformable DETR proposes using sparse sampling across key points to reduce the cost of full attention\cite{zhu2021deformable}, while Sparse DETR further sparsifies queries and keys to accelerate computation \cite{roh2021sparse}. However, less attention has been given to sparsifying the MLP layers, despite the observation that many activations in these layers are zero due to the use of ReLU, offering an untapped opportunity for computation reduction.

Beyond DETR-specific improvements, more general approaches to computational efficiency have been proposed. Mixture of Experts (MoE) models \cite{clark2022unified, geva2020transformer, shazeer2017outrageously}, originally designed to massively increase model capacity by routing inputs through large, specialized expert subnetworks, achieve computational savings by activating only a subset of the model. However, traditional MoE designs are primarily motivated by scalability rather than explicit sparsity, and they typically require training from scratch. MoE models have been scaled to thousends or even millions of fine-grained experts \cite{lepikhin2020gshard, shazeer2017outrageously, he2024mixture}.

More recently, works like DejaVu \cite{liu2023deja} attempt to predict sparsity patterns on top of pretrained vision transformers by learning lightweight modules that dynamically mask computations \cite{carion2020end}.
DejaVu offers a practical approach to sparsity prediction by introducing a small auxiliary module that forecasts which parts of a pretrained vision transformer can be safely skipped during inference.
Inspired by this line of work, we recently proposed a method that focuses on predicting sparsity specifically in attention heads, leveraging information from earlier layers of a pretrained model to make the prediction process more efficient and practical \cite{sedghiearly}. Building on these ideas, our current method seeks to exploit dynamic sparsity in pretrained DETR models with fine-grained control, without requiring extensive retraining or significant architectural modifications.

\section{Observations and Sparsity Analysis}

\begin{figure}[htb]

\begin{minipage}[b]{1.0\linewidth}
  \centering
  \centerline{\includegraphics[width=8.5cm]{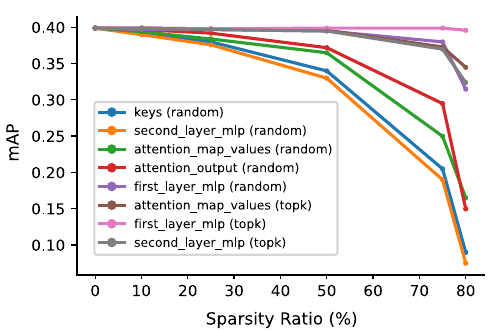}}
\end{minipage}
\caption{Activation masking in DETR using random vs. Top-K strategies across blocks. Top-K consistently reveals higher sparsity potential, highlighting the need for structured, input-aware sparsification.}
\label{fig:random}
\end{figure}

To understand the inherent sparsity and redundancy in a pretrained DETR model, we conducted an empirical analysis . We first trained a DETR model on the COCO dataset, which serves as the baseline and pretrained model throughout this work.
To explore the sparsity characteristics, we simulated artificial sparsity in two major components of the DETR architecture: the attention mechanism and the MLP (feedforward) blocks.

We applied random masking by setting a random subset of attention map values to zero in attention layers, without considering any attention magnitude or token importance (see Fig. \ref{fig:random}{}). Random masking introduces the expected level of sparsity but leads to performance degradation, indicating that naive sparsification is suboptimal. For comparison, applying a simple Top-K selection strategy, which retains only the largest attention scores, achieves much more effective sparsification with significantly less impact on performance. This result suggests that the pretrained DETR model contains considerable redundancy in the attention maps that can be structured and exploited.


Simultaneously, we analyzed the MLP blocks, which have been less studied in DETR sparsification literature. Interestingly, we observed a high degree of sparsity in the activations of the first linear layer, largely due to the ReLU activation function. However, this sparsity is highly dynamic and input-dependent, meaning that the set of inactive neurons varies significantly across different inputs. Consequently, applying random masking in the MLP blocks leads to substantial performance drops. This highlights that any effective sparsification method for MLP layers must predict activation sparsity conditioned on the input, rather than relying on fixed or random masks.

In particular, we found that a significant number of neurons are not permanently inactive ("dead"), but become zero depending on the input. This makes static pruning ineffective, yet opens the possibility for input-aware prediction of zero activations at inference time — enabling substantial computation savings if such patterns can be predicted efficiently.

\section{Computational Savings Analysis}

In this section, we analyze how our proposed gating mechanism translates activation sparsity into actual computational savings. In the original DETR architecture, the MLP  blocks include two linear layers, but only the first linear layer is followed by a ReLU activation function. Consequently, sparsity in DETR exists only at the level of activations in memory and does not directly lead to a reduction in computation, since all weights must still be processed during inference. Our method addresses this limitation by predicting the sparsity of neuron activations before the first linear layer computation occurs. By doing so, we not only avoid computing activations that would result in zeros, but also enable weight-level sparsity:

\begin{itemize}
    \item {\bf First Linear Layer}: If a neuron is predicted to be inactive (zero output), we can skip the computation associated with its weights in the first layer.

    \item {\bf Second Linear Layer}: Similarly, if the input to the second linear layer is predicted to be zero, we can skip the corresponding rows of computation.
    
\end{itemize}

This ability to predict sparsity ahead of time allows us to reduce the number of actual floating-point operations (FLOPs) required during inference in our simulations, achieving true computational acceleration rather than just memory optimization. Importantly, the amount of computation saved scales linearly with the sparsity ratio achieved by the gating mechanism. 


The extremely lightweight nature of the gating mechanism — being either a small additional layer or embedded directly within the MLP block — ensures that the overhead introduced is negligible compared to the computational benefits obtained. The overhead is approximately 10 percent of the current layer MLP FLOPs, which is negligible. This balance between minimal overhead and significant computational reduction is a core advantage of our approach.

\section{methods}


In our preliminary analysis, we observed that the first linear layer of the MLP blocks within both the encoder and decoder layers of DETR exhibits a high degree of sparsity, reaching up to 97\% on average across layers. This sparsity primarily results from the ReLU activation following the linear transformation. To explore whether this sparsity can be exploited dynamically at inference time, we designed and evaluated two different methods. The first method attempts to exploit a subset of pretrained weights to predict sparsity within the same layer. The second method involves training a small auxiliary model that learns to detect and predict sparse activations dynamically. The details of these two approaches are described in the following sections

Importantly, throughout all stages of our method, the DETR backbone remains completely frozen; no fine-tuning or retraining of the base model is performed. Our techniques introduce lightweight auxiliary modules to predict sparsity patterns dynamically during inference.

\paragraph*{Static Indicator-Based Sparsification (SIBS).}

In our first approach, we aimed to exploit the potential computational savings available in the pretrained DETR model without introducing any additional training or learnable parameters. The key idea is to identify a subset of neurons within each layer such that their activations can serve as indicators to predict the sparsity of the remaining neurons. Specifically, for each neuron $i$ in a given linear layer $
l$, we analyze the pretrained model's activations across the entire training dataset to find a subset of neurons
$S(i)$ with the following properties:

\begin{itemize}
  \item If neuron $i$'s activation is zero, then all neurons in $S(i)$ are also zero across all data points.

  \item If neuron $i$'s activation is nonzero, but the neurons in $S(i)$ are consistently zero.

\end{itemize} 

This analysis results in constructing a sparsity-predictor for each neuron. To maximize efficiency during inference, we then select a minimal set of indicator neurons whose associated subsets jointly cover as much of the layer as possible. During inference, only these indicator neurons are computed. Based on their activations and the precomputed subsets, we can dynamically infer and skip the computation of a large number of neurons that are predicted to produce zero activations, leading to significant reduction in computational cost.\\

\paragraph*{Micro-Gated Sparsification (MGS).}

Given the limited sparsity achievable by the previous static method, we developed a second approach to introduce more flexibility and adaptivity into sparsity prediction. The failure of the static method stemmed from two main observations: first, the weak correlation between the activations of different neurons, and second, the highly dynamic nature of neuron activations in response to varying input tokens. These challenges motivated the need for a learnable, dynamic mechanism.

In this method, we introduce a lightweight gating model to predict neuron activity on a fine-grained basis. The gating model consists of a single linear layer followed by a sigmoid activation function, with the number of neurons set to approximately 12\% of the size of the linear layer it aims to sparsify. Each neuron in the gating model is responsible for controlling a small subset of neurons in the original layer, acting as an importance indicator for that group. During training, the DETR model is kept frozen, and only the gating model parameters are updated. The input to the gating model is the same as the input to the original linear layer, and the training objective is formulated as a multi-label classification task. For each gate, the ground truth label is determined by computing the Euclidean norm of the corresponding subset of activations: if the norm is greater than zero, the gate's label is set to 1 (active); otherwise, it is set to 0 (inactive).

At inference time, the gating model predicts a sigmoid score for each group. If the score falls below a predefined threshold, the corresponding group of neurons is masked out, avoiding unnecessary computation. Otherwise, the full computation is performed for that group. This dynamic gating mechanism enables high levels of sparsity (up to 85–95 \%) while preserving, or even slightly improving, performance.

\section{Experiments and Results}
\subsection{Static Indicator-Based Sparsification (SIBS)}
As described in Section 5, the first method we explored was the Static Indicator-Based Sparsification (SIBS) approach. For each neuron in the first linear layer of the MLP blocks, we identified a subset of other neurons whose activation sparsity could be inferred based on the activation state (zero or nonzero) of the indicator neuron. This was constructed heuristically by analyzing activations across the training set and was stored as a lookup table for use during inference.

For our experiments, we trained a DETR model on the COCO 2017 train split to serve as the baseline pretrained model. Since the SIBS method is purely rule-based and relies on heuristic rules applied to pretrained weights and activations, no additional retraining was necessary. The heuristic subset construction was conducted over a single epoch of the training data to capture activation patterns. Importantly, as this method does not modify model weights or introduce any learnable parameters, the performance (e.g., object detection mAP) of the model remains unchanged after applying sparsification.
During inference, the process involves computing activations only for a fixed set of indicator neurons per layer. Based on the computed values and the precomputed lookup dictionary, we dynamically mask the remaining neurons before proceeding with full computation. This allows selective computation of only the necessary parts of the MLP block.

We conducted experiments across different configurations, varying the size of the indicator neuron set. Specifically, we tested: 16 indicators (approximately 1\% of the layer size),
256 indicators (approximately 12, and
    512 indicators (approximately 25\%).
As shown, even with larger indicator sets, the number of neurons that could be predicted and masked does not increase significantly.

\begin{table}[ht]
\centering
\begin{tabular}{|c|c|c|c|}
\hline
\textbf{Indicator Set Size} & \textbf{Covered Nodes} & \textbf{Sparsity (\%)} \\
\hline
15 & 351 & 17 \\
\hline
256 & 669 & 20 \\
\hline
512 & 672 & 20 \\
\hline
\end{tabular}
\caption{Performance across different indicator set sizes, showing covered nodes and sparsity.}
\label{tab:indicator_performance}
\end{table} 
These findings indicate that the sparsity achievable with this rule-based method is relatively limited, reaching less than a 300\% "sparsity amplification" (where 100\% would correspond to one predicted node per indicator) (see table \ref{tab:indicator_performance}).
The relatively flat performance across different indicator set sizes suggests that much of the detectable sparsity corresponds to permanently inactive ("dead") neurons rather than dynamic activation patterns.
Moreover, the overall FLOPs reduction at the MLP block level remained minor, reflecting the fact that rule-based prediction cannot fully exploit the dynamic nature of activation sparsity.
This analysis highlights a critical limitation of the static heuristic method: the remaining active neurons exhibit low correlation and high variability depending on the input data. As a result, purely deterministic sparsification cannot capture the full dynamic sparsity potential of the model, motivating the development of our second method based on learned dynamic gating.

\subsection{Micro-Gated Sparsification (MGS) Evaluation}
For the Micro-Gated Sparsification (MGS) approach, we trained a lightweight gating mechanism consisting of a single linear layer with 256 neurons (approximately 12\% of the size of the original MLP layer), followed by a sigmoid activation function. The gating model was trained on top of the frozen pretrained DETR model using the COCO 2017, following the method described in Section 3. Each gate was trained in a multi-label classification fashion, where the ground truth was determined by the Euclidean norm of the corresponding neuron group in the original layer.
The gating model was trained for 5 epochs, and evaluation was performed on the COCO 2017 validation split. During inference, a threshold was applied to the sigmoid outputs to determine whether to activate or mask each neuron group dynamically.
We evaluated the MGS method under three different configurations:

\begin{table}[ht]
\centering
\begin{tabular}{|c|c|c|c|c|}
\hline
\textbf{Layer} & \textbf{Acc (\%)} & \textbf{mAP} & \textbf{Spar. (\%)} & \textbf{Threshold} \\
\hline
0  & 86.80 & \cellcolor[HTML]{F8696b}0.384 & 94.20 & 0.5 \\
1  & 85.13 & \cellcolor[HTML]{f98788}0.387 & 94.11 & 0.5 \\
2  & 80.68 & \cellcolor[HTML]{f99192}0.388 & 86.20 & 0.5 \\
3  & 79.74 & \cellcolor[HTML]{fcc3c3}0.393 & 74.36 & 0.5 \\
4  & 86.73 & \cellcolor[HTML]{fef5f5}0.398 & 86.10 & 0.5 \\
5  & 92.09 & \cellcolor[HTML]{ffffff}0.399 & 88.96 & 0.5 \\
6  & 85.31 & \cellcolor[HTML]{c4dcf3}0.400 & 90.23 & 0.5 \\
7  & 87.04 & \cellcolor[HTML]{fef5f5}0.398 & 94.27 & 0.5 \\
8  & 88.06 & \cellcolor[HTML]{f98788}0.387 & 94.99 & 0.5 \\
9  & 91.77 & \cellcolor[HTML]{4d93d9}0.402 & 96.90 & 0.5 \\
10 & 94.83 & \cellcolor[HTML]{89b8e6}0.401 & 98.23 & 0.5 \\
11 & 96.70 & \cellcolor[HTML]{c4dcf3}0.400 & 98.35 & 0.5 \\
\hline
vanilla & 0.0 & 0.399 & 0.0 & 0.0 \\
\hline
all & 96.70 & 0.329 & 91.42 & 0.5 \\
\hline
\end{tabular}
\caption{Per-layer performance with gate accuracy, mAP, sparsity, and fixed threshold.}
\label{tab:layer_performance}
\end{table}

\paragraph*{Fixed Threshold (0.5 Across All Layers).}

In the first setting, we applied a fixed threshold of 0.5 to the sigmoid outputs across all layers. The results are summarized in Table \ref{tab:layer_performance}, which includes, the impact of gating on object detection performance (measured by mean Average Precision, mAP), the gating accuracy (i.e., accuracy of sparsity prediction per layer), and the achieved sparsity ratio after applying dynamic masking.
Surprisingly, despite applying aggressive sparsification, the model showed only a slight reduction in overall mAP. Furthermore, layer-by-layer ablation revealed that in some cases, sparsification improved performance: masking out redundant or noisy activations in certain layers actually led to a boost of up to +2\% mAP. This suggests that dynamic gating can also serve as a form of regularization by filtering out unhelpful computations.
Overall, high sparsity ratios (up to 85–95\%) were achieved across many layers, validating the effectiveness of our fine-grained gating approach.

\begin{table}[ht]
\centering

\begin{tabular}{|c|c|c|c|c|}
\hline
\textbf{Layer} & \textbf{Acc (\%)} & \textbf{mAP} & \textbf{Spar. (\%)} & \textbf{Threshold} \\
\hline
0  & 86.80 & \cellcolor[HTML]{f8696b}0.389 & 64.55 & 0.1 \\
1  & 85.13 & \cellcolor[HTML]{ffffff}0.399 & 71.02 & 0.1 \\
2  & 80.68 & \cellcolor[HTML]{fde1e1}0.397 & 43.20 & 0.1 \\
3  & 79.74 & \cellcolor[HTML]{c4dcf3}0.400 & 74.36 & 0.1 \\
4  & 86.73 & \cellcolor[HTML]{fef0f0}0.398 & 86.10 & 0.5 \\
5  & 92.09 & \cellcolor[HTML]{ffffff}0.399 & 88.96 & 0.5 \\
6  & 85.31 & \cellcolor[HTML]{c4dcf3}0.400 & 90.23 & 0.5 \\
7  & 87.04 & \cellcolor[HTML]{fef0f0}0.398 & 94.27 & 0.5 \\
8  & 88.06 & \cellcolor[HTML]{fbfbfb}0.399 & 73.71 & 0.25 \\
9  & 91.77 & \cellcolor[HTML]{4d93d9}0.402 & 96.90 & 0.5 \\
10 & 94.83 & \cellcolor[HTML]{89b8e6}0.401 & 98.23 & 0.5 \\
11 & 96.70 & \cellcolor[HTML]{c4dcf3}0.400 & 98.35 & 0.5 \\
\hline
vanilla & 0.0 & 0.399 & 0.0 & 0.0 \\
\hline
all & 96.70 & 0.394 & 81.65 & dynmic \\
\hline
\end{tabular}
\caption{Alternative dynamic threshold configurations per layer to improve mAP while maintaining high sparsity. Thresholds were relaxed for select layers to better balance performance and efficiency.}
\label{tab:alt_layer_performance}
\end{table}

\paragraph*{Flexible Threshold Adjustment.}

Motivated by the layer-specific behavior observed, we introduced a flexible thresholding strategy. In this second setting, we manually reduced the threshold for layers where aggressive sparsification led to significant performance drops, making the gating less strict for critical layers while maintaining tight sparsity for others. The adjusted thresholds led to better mAP results, with performance closer to the baseline DETR model while still achieving substantial sparsity, as shown in Table 4.
This experiment highlights the importance of adapting the aggressiveness of sparsity on a per-layer basis to optimize the trade-off between computation savings and accuracy.

\begin{table}[ht]
\centering

\begin{tabular}{|c|c|c|c|c|}
\hline
\textbf{Layer} & \textbf{Acc (\%)} & \textbf{mAP} & \textbf{Spar. (\%)} & \textbf{Threshold} \\
\hline
0  & 86.80 & \cellcolor[HTML]{ffffff}0.399 & 0.00  & 0    \\
1  & 85.13 & \cellcolor[HTML]{ffffff}0.399 & 71.02 & 0.1  \\
2  & 80.68 & \cellcolor[HTML]{ffffff}0.399 & 0.00  & 0    \\
3  & 79.74 & \cellcolor[HTML]{c4dbf3}0.400 & 74.36 & 0.1  \\
4  & 86.73 & \cellcolor[HTML]{ffffff}0.399 & 72.14 & 0.25 \\
5  & 92.09 & \cellcolor[HTML]{ffffff}0.399 & 88.96 & 0.5  \\
6  & 85.31 & \cellcolor[HTML]{c4dbf3}0.400 & 90.23 & 0.5  \\
7  & 87.04 & \cellcolor[HTML]{fcfcfc}0.399 & 72.82 & 0.25 \\
8  & 88.06 & \cellcolor[HTML]{ffffff}0.399 & 73.71 & 0.25 \\
9  & 91.77 & \cellcolor[HTML]{4d93d9}0.402 & 96.90 & 0.5  \\
10 & 94.83 & \cellcolor[HTML]{89b7e6}0.401 & 98.23 & 0.5  \\
11 & 96.70 & \cellcolor[HTML]{c4dbf3}0.400 & 98.35 & 0.5  \\
\hline
vanilla & 0.0 & 0.399 & 0.0 & 0.0 \\
\hline
all & 96.70 & 0.399 & 69.72 & dynmic \\
\hline
\end{tabular}
\label{tab:zero_thresh_config}
\caption{Alternative dynamic threshold configurations per layer to improve mAP while maintaining high sparsity. Thresholds were relaxed for select layers to best performance}
\end{table}

\paragraph*{Threshold Tuning for Baseline Matching.}

\begin{figure}[htb]

\begin{minipage}[b]{1.0\linewidth}
  \centering
  \centerline{\includegraphics[width=8.5cm]{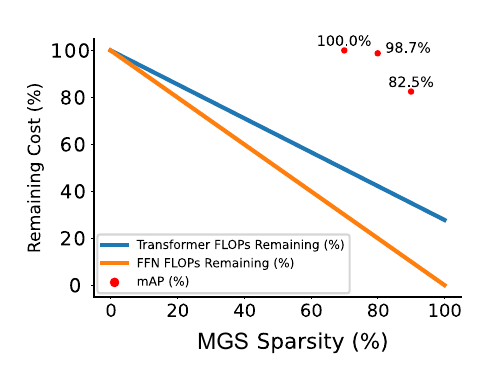}}
\end{minipage}
\caption{FLOPS reduction}
\label{fig:flops reduction}
\end{figure}

In the final setting, we further tuned thresholds layer-by-layer to match the baseline DETR performance as closely as possible. This process involved significantly lowering the threshold for certain sensitive layers, effectively disabling gating for them when necessary. While this resulted in a reduction of overall sparsity (down to around 60\%), we still observed high sparsity in many layers and considerable FLOPs reduction overall. Fig.\ref{fig:flops reduction} shows the trade-off between achieved sparsity, FLOPs savings, and detection performance for all configurations.

\section{Discussion}


These experiments demonstrate that our MGS method enables a flexible, dynamic, and fine-grained control of sparsity at inference time. Unlike static MoE-like models, MGS adapts to input complexity, allowing easier samples to consume less computation and harder samples to retain sufficient capacity. The dynamic thresholding strategy offers an effective mechanism for navigating the trade-off between computational efficiency and model accuracy, potentially enabling greater savings on simpler inputs without compromising performance.



\bibliographystyle{IEEEbib}
\bibliography{mlsp_template_camera_ready}

\end{document}